\documentclass{article}

\PassOptionsToPackage{numbers, compress}{natbib}

\usepackage[sglblindworkshop, final]{ai4nextg_neurips_2025}

\usepackage[utf8]{inputenc} 
\usepackage[T1]{fontenc}    
\usepackage{hyperref}       
\usepackage{url}            
\usepackage{booktabs}       
\usepackage{amsfonts}       
\usepackage{nicefrac}       
\usepackage{microtype}      
\usepackage{xcolor}         

\usepackage{graphicx}
\usepackage{caption}

\title{A Probabilistic Approach to Pose Synchronization for Multi-Reference Alignment with Applications to MIMO Wireless Communication Systems}

\usepackage{amsmath,amsfonts,bm}
\usepackage{cancel}

\def\eqref#1{eq.~\ref{#1}}

\def\1{\bm{1}}

\def\mZero{{\bm{0}}}

\def\mA{{\bm{A}}}
\def\mB{{\bm{B}}}

\def\mH{{\bm{H}}}
\def\mI{{\bm{I}}}

\def\mM{{\bm{M}}}

\def\mP{{\bm{P}}}

\def\mR{{\bm{R}}}

\def\mU{{\bm{U}}}
\def\mV{{\bm{V}}}
\def\mW{{\bm{W}}}
\def\mX{{\bm{X}}}
\def\mY{{\bm{Y}}}

\def\mmu{{\bm{\mu}}}
\def\mSigma{{\bm{\Sigma}}}

\DeclareMathAlphabet{\mathsfit}{\encodingdefault}{\sfdefault}{m}{sl}
\SetMathAlphabet{\mathsfit}{bold}{\encodingdefault}{\sfdefault}{bx}{n}

\def\gN{{\mathcal{N}}}

\usepackage{amsthm}
\usepackage{amssymb}

\theoremstyle{definition}

\theoremstyle{remark}

\newcommand*{\prob}[1]{\mathbb{P}}

\newcommand{\E}{\mathbb{E}}

\DeclareMathOperator*{\argmax}{arg\,max}

\DeclareMathOperator{\Tr}{Tr}

\renewcommand{\vec}[1]{\ensuremath{\operatorname{vec}\left({#1}\right)}}

\usepackage{centernot}

\author{%
  Rob Romijnders \\
  University of Amsterdam\\
  QUvA-Lab\\
  \And
  Gabriele Cesa,
  Christos Louizos,
  Kumar Pratik,
  Arash Behboodi \\
      Qualcomm AI Research, Amsterdam
}

\begin{document}

\maketitle

\begin{abstract}
From molecular imaging to wireless communications, the ability to align and reconstruct signals from multiple misaligned observations is crucial for system performance.
We study the problem of multi-reference alignment (MRA), which arises in many real world problems, such as cryo-EM, computer vision and, in particular, wireless communication systems.
Using a probabilistic approach to model MRA, we find a new algorithm that uses relative poses as nuisance variables to marginalize out -- thereby removing the global symmetries of the problem and allowing for more direct solutions and improved convergence.
The decentralization of this approach enables significant computational savings by avoiding the cubic scaling of centralized methods through cycle consistency. Both proposed algorithms achieve lower reconstruction error across experimental settings.
\end{abstract}

\section{Introduction}

Multi-reference alignment (MRA) is a fundamental problem in signal processing and statistical inference that arises when reconstructing a signal from multiple noisy observations, and each observation has been transformed by an unknown group action such as rotation, translation, or other geometric transformation. This problem appears in diverse applications across scientific domains, from cryo-electron microscopy (Cryo-EM) \cite{nogales_development_2016,henderson_model_1990} and cryo-electron tomography (Cryo-ET) for molecular structure determination~\cite{singer2020computational,singer2018mathematics,cryo_em}, to computer vision applications like Structure-from-Motion (SfM) and Simultaneous Localization and Mapping (SLAM), and even narrow-band 5G NR~\cite{requestnet}.

A core challenge in MRA lies in the fact that unknown transformations of the signal create an ambiguity. The signal could be recovered if the poses were known, and the unknown poses could be recovered when the signal was known. Moreover, the original signal can only be recovered up to a global transformation 
and, likewise, the observations' misalignments (i.e., their absolute poses) are only defined up to a joint transformation.
In other words, any global transformation applied to all observations' poses yields an equally valid solution. 
Traditional approaches to MRA typically treat the absolute poses of the observations as nuisance variables to be estimated or marginalized over in an intermediate step before signal reconstruction. This includes various probabilistic approaches based on variants of the expectation-maximization algorithm~\cite{cryo_em} or other message-passing algorithms~\cite{perry2018message}. However, these methods can face challenges due to the inherent global symmetry of the problem.

Indeed, any disjoint estimate of the absolute poses necessarily breaks the global symmetry of the problem, which can lead to optimization difficulties in expectation-maximization or maximization-maximization algorithms. The fundamental issue is that global poses are not uniquely defined. Any global transformation of the poses leads to another equally valid assignment. This can create convergence issues or trap optimization algorithms in local minima, e.g., due to subsets of observations initially converging to different global alignments.

In this work, we propose a novel probabilistic approach to MRA that addresses these challenges by reformulating the problem in terms of relative poses rather than absolute ones. 
Our key insight is that relative poses, in contrast to absolute ones, are uniquely defined and do not suffer from the global symmetries that can cause optimization difficulties.
However, not all possible assignments of relative poses correspond to a set of absolute ones.
As we prove in Appendix~\ref{app:triple_synchronization_proof}, relative poses that are cycle-consistent are in one-to-one correspondence with equivalence classes of absolute pose assignments.
Hence, to ensure that the relative poses represent valid configurations, we incorporate a cycle consistency constraint into our probabilistic model, which enforces that walks along any cycle in the relative-poses graph lead to identity transformations.

As a concrete example of the multi-reference alignment problem, we consider the problem of channel estimation (CE) in narrow-band 5G NR wireless communication systems~\cite{channels2020modulation, requestnet}. In the multi-input multi-output (MIMO) setting, the 5G NR specification allows channels to be precoded individually. Still, signals in neighboring PRG blocks can be correlated (Orthogonal Frequency Modulation and Demodulation). Current receivers do not utilize this correlation, resulting in suboptimal performance. We are the first to combine the group-theoretic approaches from multi-reference alignment with the application of 5G decoding using statistical methods, demonstrating how our relative pose framework can be applied to real-world communication systems.

Finally, we propose different algorithms to efficiently perform inference on this graph and evaluate them on multiple synthetic datasets, showing significant improvements over traditional approaches that rely on absolute pose estimation.
In summary, this paper makes the following key contributions:
\begin{itemize}
\item We present the first MIMO decoding framework that explicitly incorporates relative rotations, addressing a fundamental limitation of existing methods.
\item We develop both a direct method and an iterative refinement algorithm for joint denoising and synchronization, grounded in a principled probabilistic model, also known as a structured probabilistic graphical model.
\item We provide theoretical insights and empirical evidence of the advantages of our approach in 5G-like MIMO scenarios.
\end{itemize}

The remainder of this paper is organized as follows: Section~\ref{sec:problemformulation} formulates the problem, introduces our probabilistic graphical model, and reviews related work on MIMO synchronization and pose estimation. Section~\ref{sec:method} details the proposed algorithms. Section~\ref{sec:results} presents experimental results, and Section~\ref{sec:conclusion} concludes the paper with directions for future research.

\vspace{-1mm}
\section{Problem formulation}\label{sec:problemformulation}
\vspace{-1mm}

\paragraph{Background on Narrow-band 5G NR MIMO system} 
The 5G NR communication system uses an orthogonal-frequency division multiplexing (OFDM) scheme to communicate messages along orthogonal subcarrier signals at each time step.
As a result, the communication channel is divided into a time-frequency grid, with each cell (or resource element, RE) carrying a single symbol.
In a multiple-input-multiple-output (MIMO) scheme, both the transmitter and receiver have multiple antennas, which enables the transmission of multiple symbols on the same time-frequency location.
Formally, assuming $d$ transmitting and receiving antennas, the transmission at a frequency $f$ and time step $t$ of a symbol vector $x_{f,t} \in \mathbb{C}^d$ is 
\begin{align}
    y_{f,t} = \mH_{f, t} \mP_{f, t} x_{f, t} + \varepsilon
\end{align}
where $H_{f, t} \in \mathbb{C}^{d \times d}$ is the MIMO channel matrix, $y \in \mathbb{C}^d$ the received signal, $\varepsilon \sim \mathcal{N}_\mathbb{C}(0, \sigma^2I)$ is Gaussian noise and $P\in U(n)$ is a unitary matrix which defines the \emph{precoding}. 
The transmitter precodes the transmitted messages to help minimize interference and improve SNR, e.g., by aligning the message along the singular vectors of the channel matrix.
For practical reasons, in narrow-band 5G NR, the precoding matrix is shared by multiple neighboring resource elements within the same physical resource block group (PRG).
The challenge for the receiver is estimating the effective channel matrices (i.e., after precoding) from a few noisy observations obtained via some pilot messages. We consider a multi-reference alignment problem inspired by this narrow-band 5G NR channel estimation problem.
Our problem formulation allows us to abstract away from many technical details of the communication protocol, while preserving its relevant properties and the underlying synchronization challenges. For more details, we refer to~\citet{5g_fundamentals,requestnet} for a more precise description of the real channel estimation problem in narrow-band 5G NR.

\paragraph{Abstract formulation}
For ease of modelling, we model the channel matrix as real numbers and the precoding matrices as real orthogonal matrices $\mP \in O(d) \subset \mathbb{R}^{d \times d}$. 
We denote the channel in the $j$-th PRG block by $\mH_i\in\mathbb{R}^{D \times d}$, where the dimension $D$ indexes all time-frequency positions within the block and the output antennas.
In a realistic 5G MIMO scenario, the receiver observes the effective channels $\mH_i^{'}=\mH_i \mP_i$ via some pilot messages only at certain time-frequency positions (demodulation reference signal, DMRS). This signal is then used to estimate the full channel via linear predictors.
We assume the receiver observes the full channel\footnote{Note that pilot DMRS signals provide linear measurements of the channel, so our simplification reduces to the use of a diagonal full rank covariance matrix between $\mB_i$ and $\mH^{'}$, rather than a more generic one.} corrupted by real Gaussian noise, i.e. $\mB_i = \mH_i \mP_i + \varepsilon_i \in \mathbb{R}^{D \times d}$ with $\varepsilon_i \sim \mathcal{N}(0, \sigma^2_\varepsilon I)$.

Taken all random variables together, $\mH$, $\mP$, and $\mB$, the joint probability distribution is:
\begin{equation}\label{eqn:joint_distr}
p(\mB, \mP, \mH) = p(\mB|\mH^{'}) \ p(\mH^{'} | \mH,\mP)  \ p(\mP) \ p(\mH)
\end{equation}
where $ p(\mB|\mH^{'}) = \mathcal{N}(\mB | \mH^{'}, \sigma^2 I)$, $p(\mH^{'} | \mH,\mP) = \delta_{\mH^{'}-\mH\mP}$ is a deterministic factor applying the precoding to the channel.
We assume the prior $p(\mP)$ on the rotation precoding matrices is uniform over the $O(d)$ group.
Finally, $p(\mH)$ specifies a rotation invariant\footnote{This is a reasonable assumption since it is equivalent to assuming zero mean and that the correlation depends only on the frequency and time position and is the same among all the antennas, since the precoding transformation only acts on the antennas' dimension of the channels matrices.} prior via a matrix normal distribution over all  channel matrices $\mathcal{MN}(0, \mU, \mI)$ with $\mI$ the identity matrix and matrix $\mU$ modelling the row-covariance. 
For the rotation invariance, if $P\in O(d)$ and $\mX \sim \mathcal{MN}(0, \mU, \mI)$ then $\mX\mP \sim \mathcal{MN}(0, \mU, \mP^T\mP=\mI)$.
We also refer to this prior as `spatial correlation,' indicating correlation along the time and frequency axes among the PRG blocks due to Rayleigh scattering~\cite{rayleigh_fading,5g_fundamentals}.

This spatial correlation enables the estimation of the channel in a PRG block $\mH_i^{'}$ from its noisy observations $\mB_i$.
However, the unknown precoding matrices prevent the receiver from leveraging the correlation between neighboring blocks.
Our key insight is that by modeling the relative rotations $\mR_{ij} = \mP_i^{-1}\mP_j$ between adjacent blocks in the time-frequency grid, we can leverage the spatial and temporal correlations to achieve more robust channel estimation.

\textbf{Absolute to relative poses}
Multi-reference alignment problems typically present a global symmetry, which implies that absolute poses and the underlying signal are not uniquely defined.
Indeed, the set of channels $\mH=\{\mH_i\}_i$ and precoding matrices $\mP=\{\mP_i\}_i$ generate the same observations $\mB=\{\mB_i\}_i$ as the channels $\bar{\mH}=\{\mH_i Q\}_i$ and the precoding $\bar{\mP}=\{Q^{-1} \mP_i\}_i$ for any transformation $Q \in O(d)$.
Any attempt to estimate these quantities requires a form of symmetry breaking.
However, this problem vanishes with a local viewpoint and, therefore, considering relative poses; see Appendix~\ref{app:absolute_to_relative}
For this reason, we develop our method to recover relative poses $\mR = \{\mR_{i,j} := \mP_i^{-1}\mP_j \}_{i, j}$, which are uniquely defined and suitable for an efficient distributed and local computational scheme.

\textbf{Relation to prior work in synchronization: } 
synchronization over compact groups has been approached with approximate message passing methods \cite{perry2018message} or semi-definite programming~\cite{bandeira2017estimation}. Other works studied CryoEM~\cite{cryo_em} and stochastic block models~\cite{DBLP:journals/jmlr/Abbe17}.
The fundamental difference in our work, though, is that we aim to estimate the signal itself. \cite{perry2018message} assume noisy observation of the relative poses and strive to find an optimal representation of the absolute poses. In our case, we have noisy observations of the underlying signal at each location. The goal is to efficiently and effectively marginalize over the relative poses to obtain a better estimate of the channel. Although our approach utilizes similar estimations of the relative poses, it does not synchronize the absolute poses. Instead, it can be said informally that we `directly synchronize' the underlying signal.
\cite{requestnet} first describes the MRA of narrow-band 5G NR channel estimation, but adopted complex deep learning architectures.
\textbf{Relation to wider related work} Synchronization problems have been studied in a more abstract context for their group-theoretic properties~\cite{sync_cohomology_01,sync_cohomology_02}. Some works have studied pairwise relative pose estimation by candidate elimination~\cite{Sun_2023_CVPR}, but we are the first structured statistical model to marginalize over the unknown poses to reconstruct the underlying signal effectively and efficiently.

\section{Method}

\subsection{Direct estimation}

A direct approach is to estimate the relative rotations $\mR_{ij}$ between adjacent nodes $\mH$ in the time-frequency grid. This is done by maximizing the likelihood of the data $\log p(\mB)$. Observations $\mB \in \mathbb{R}^{D \times d}$ is the matrix of $D$ times a $d$ dimensional observation. However, the direct approach has two disadvantages: estimating in pairs does not enable global synchronization, and direct estimates can be suboptimal as rotation and observation noise are confounded, where estimating one changes the other. We address both properties in the next section.

Estimating the channel $\mH^{'}$ can be done by writing the posterior given the data $\mB$ directly:
\begin{align}
\argmax_{\mH^{'}}  p(\mH^{'} | \mB) &= \int_\mR  p(\mH^{'} | \mB, \mR) p(\mR | \mB) d\mR
\end{align}

The posterior $p(\mR | \mB)$ can be heuristically approximated with a delta distribution -- meaning that we can take a point estimate of the rotation. 
Assuming only two blocks $\{\mH_1, \mH_2\}$, if the covariance matrix of the prior on the correlated channel is $\mU$, then one can find the relative rotation $\mR_{12}$ between the two blocks from the following projection: $\argmax_{\mR_{12}}  \langle  \mR_{21}, \mH^{'T}_2  \mU_a^T \mH^{'}_1 \rangle_F$.

Likewise, for a triplet of three nodes $\mH_1, \mH_2, \mH_3$ and relative poses $\mR_{12}$ and $\mR_{13}$ (note that $\mR_{23} := \mR_{12}^T \mR_{13}$ and that $\mR_{ij} = \mR_{ji}^T$), one can iterate the two updates below in Equation~\ref{eq:update_triplet} until convergence. Matrices $\mU_a, \mU_b, \cdots$ are submatrices of the covariance matrix $\mU$, as described with Equation~\ref{eqn:subslice_mU} in the Appendix.
\begin{equation}\label{eq:update_triplet}
\begin{cases}
\mR_{13} = \argmax \  \textcolor{gray}{\langle  \mR_{21}, \mB_2^T  \mU_a^T \mB_1 \rangle_F} + \langle  \mR_{31}, \mB_3^T  \mU_b^T \mB_1 \rangle_F  + \langle  \mR_{31}, \mB_3^T  \mU_c^T \mB_2 \mR_{21} \rangle_F  \\
\mR_{12} = \argmax \   \langle  \mR_{21}, \mB_2^T  \mU_a^T \mB_1 \rangle_F + \textcolor{gray}{\langle  \mR_{31}, \mB_3^T  \mU_b^T \mB_1 \rangle_F } + \langle  \mR_{21}, \mB_2^T  \mU_c \mB_3 \mR_{31} \rangle_F.
\end{cases}
\end{equation}
The derivation is outlined in Appendix~\ref{app:direct3_derivation}. Crucially, the above solution arises from using properties of the Matrix Normal distribution. Ending up with the above Frobenius inner-products is a result of the Orthogonal Procrustes problem~\cite{procrustes01,procrustes02}, which is a mean-square solution when optimizing over the set of unitary matrices. The projection is a closed-form maximizer of the Procrustes problem. Therefore, each step of the iteration is non-decreasing in the objective value, and we find that this usually converges in three to five steps.

Secondly, obtaining the channel estimate from the observations and rotation matrices $\argmax p(\mH^{'} | \mB, \mR)$ is a linear problem, which can be solved in $O(D^3)$ time:

\begin{align}\label{eqn:mstep_direct}
\begin{bmatrix} \mH^{'}_{1} \\ \mH^{'}_{2}\hat{\mR}_{21} \\ \mH^{'}_{3}\hat{\mR}_{31}  \end{bmatrix} = \Big(\sigma^{2}_\epsilon \mU^{-1} + \mI_{3D}\Big)^{-1} \begin{bmatrix} \mB_1 \\ \mB_2 \hat{\mR}_{21} \\ \mB_3 \hat{\mR}_{31}  \end{bmatrix} 
\end{align}

The above solutions are worked out for a `triplet` of nodes, meaning that we consider three nodes and derive the estimation equations. This could be worked out for an arbitrary number of nodes. However, this has two downsides: (a) the computation scales cubically in the number of measurements and thus cubically in the number of nodes. 
Hence, scaling this approach on the wide time-frequency lattices of up to hundreds blocks per time slot is impractical, especially under the resource and power constraints of modern 5G telecom modems~\cite{5g_fundamentals}. 
The second downside (b) to including more nodes is that synchronization becomes `harder.' After all, only noisy nodes are available, and harmonizing among more noisy observations pertaining to random rotations has more numerical issues. 

Therefore, we suggest decentralizing the estimation into triplets. A triplet of nodes is the smallest subgraph in which a local synchronization can imply a global synchronization over the entire grid, see Appendix~\ref{app:triple_synchronization_proof}. Meanwhile, a triplet of nodes is most economical with the local `spatial correlation' that exists among the channels. Due to Rayleigh scattering~\cite{rayleigh_fading}, only nearby PRG blocks will have correlation, and denoising among local subgraphs most economically uses the compute resources.

\subsection{Synchronization and denoising via iterative refinement}\label{sec:method}

From the direct solution, we suggest an iterative refinement method that is loosely inspired by the Expectation Maximization algorithm. The direct solution alternates between two maximizations, over $p(\mH^{'} | \mB, \mR)$ and over $p(\mR | \mB)$. The latter maximization assumes that channel observations $\mB$ are made and the relative poses can be estimated from them. By extension, $\mH^{'}$ will serve as a `denoised version' of the channel. As such, we will report the Mean-Square error between $\mH^{'}$ and $\mB$. Then, as $\mH^{'}$ is a denoised version of $\mB$, we can refine the rotation estimate from $p(\mR | \mH^{'})$ instead. Continuing, the relative poses can be found from $p(\mH^{'} | \mB, \mR)$.

Loosely, this is inspired by the Expectation-Maximization algorithm~\cite{bishop,murphy}. Analogous to the M-step, we estimate the parameters of the channel using an estimate for the rotation random variable. Analogous to the E-step, we find an estimate for the relative poses. In this case, however, we take only a point estimate, $q(\mR) = \delta_\mR $ in $\log p(\mB) \geq \max_{q(\mR)}   \E_{\mR \sim q(\mR)}[\log p(\mB, \mR, \mH^{'}_t) - \log q(\mR)]$.

The estimation for the relative poses can be done using similar Frobenius projections, starting from $\argmax_\mR p(\mR | \mH^{'})$. This time, however, the estimates for the channel are used:
\small
\begin{equation}
\begin{cases}
\mR_{12} = \argmax_{\mR_{12}}  \langle  \mR_{21}, \mH^{'T}_{2,t}  \mU_a^T \mH^{'}_{1,t} \rangle_F + \textcolor{gray}{\langle  \mR_{31}, \mH^{'T}_{3,t}  \mU_b^T \mH^{'}_{1,t} \rangle_F}  + \langle  \mR_{21}, \mH^{'T}_{2,t}  \mU_c \mH^{'}_{3,t} \mR_{31} \rangle_F  \\
\mR_{13} = \argmax_{\mR_{13}}  \textcolor{gray}{\langle  \mR_{21}, \mH^{'T}_{2,t}  \mU_a^T \mH^{'}_{1,t} \rangle_F} + \langle  \mR_{31}, \mH^{'T}_{3,t}  \mU_b^T \mH^{'}_{1,t} \rangle_F  + \langle  \mR_{31}, \mH^{'T}_{3,t}  \mU_c^T \mH^{'}_{2,t} \mR_{21} \rangle_F \label{eqn:iter_R}
\end{cases}
\end{equation}
\normalsize

The channel estimation is derived in Appendix~\ref{app:iterative3_derivation} and results in an estimator similar to Equation~\ref{eqn:mstep_direct}:

\begin{equation}\label{eqn:iter_H}
\begin{bmatrix} \mH^{'}_{1, t+1} \\ \mH^{'}_{2, t+1}\hat{\mR}_{21} \\ \mH^{'}_{3, t+1}\hat{\mR}_{31}  \end{bmatrix} = \Big(\sigma^{2}_\epsilon \mU^{-1} + \mI_{3D}\Big)^{-1} \begin{bmatrix} \mH^{'}_{1,t} \\ \mH^{'}_{2,t} \hat{\mR}_{21} \\ \mH^{'}_{3,t} \hat{\mR}_{31}  \end{bmatrix}
\end{equation}

Together, Equations~\ref{eqn:iter_R} and~\ref{eqn:iter_H} can be repeated to refine the channel estimate and the rotation estimate. We typically find that this stabilizes after three or four iterations, which is illustrated in the experimental results section.

\begin{figure}
    \centering
    \includegraphics[width=\linewidth]{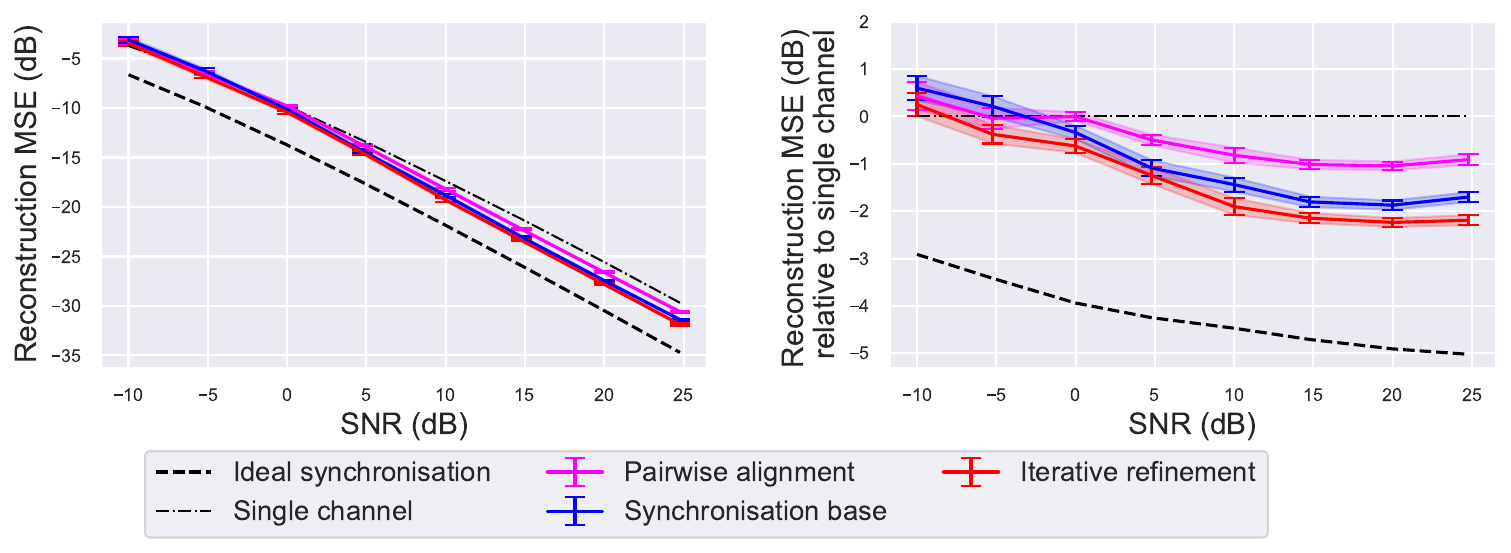}
    \caption{Iterative refinement achieves the lowest reconstruction error compared to signal denoising only. This is a synthetic 6-by-6 grid of 12 measurements each. Error bars indicate standard error among 25 random seeds. Additionally, the `synchronization base' approach, by itself, achieves lower reconstruction error than prior art for reasonable signal-to-noise ratios.}
    \label{fig:main_sidebyside}
\end{figure}

\section{Experimental results} \label{sec:results}
\vspace{-1mm}

\begin{figure}[t]
\centering
\begin{minipage}[t]{0.48\textwidth}
\centering \vspace{-0.01\baselineskip}
\includegraphics[width=\linewidth]{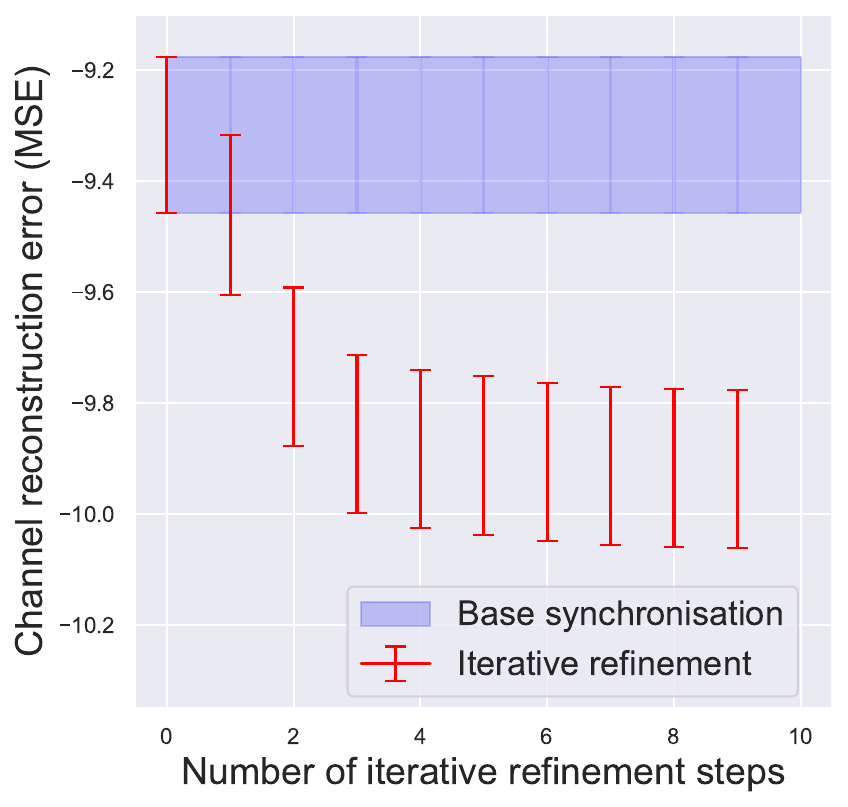}
\captionof{figure}{The reconstruction error for varying number of iterative refinement steps. Most improvement over the direct approach (blue shaded region) is achieved in two to five steps.}
\label{fig:left}
\end{minipage}\hfill
\begin{minipage}[t]{0.48\textwidth}
\centering \vspace{-0.01\baselineskip}
\includegraphics[width=\linewidth]{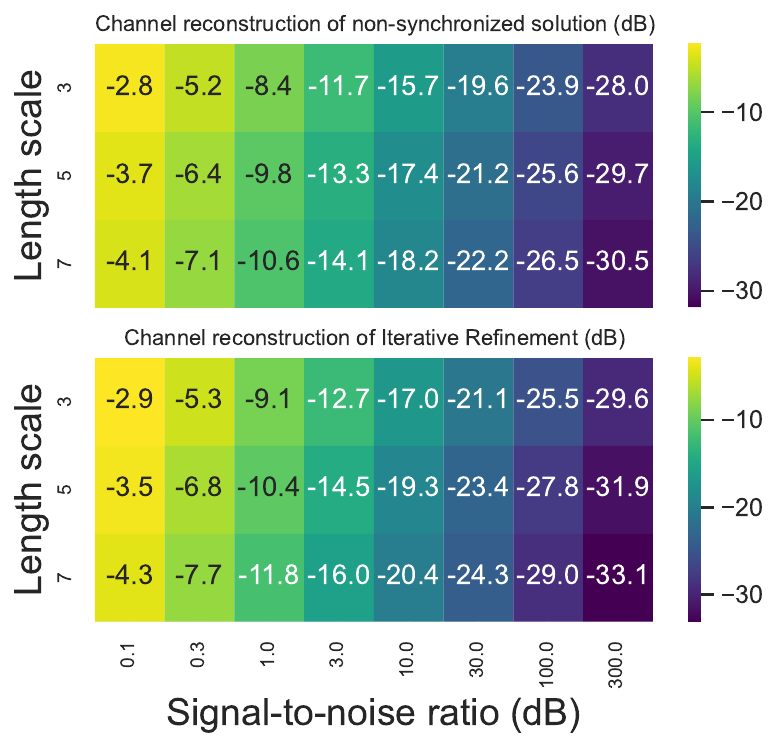}
\captionof{figure}{Numerical results along varying length scales, which determine the correlation properties. Our synchronization improves the channel reconstruction across these settings.}
\label{fig:heatmap_lengthscale_snr}
\vspace{-5mm}
\end{minipage}
\end{figure}

Our experimental evaluation demonstrates the advantages of our proposed approach. We compare three methods. Starting with the `pairwise alignment' that operates on pairs of nodes in the time-frequency grid. It serves as the baseline, and, while computationally efficient, it suffers from the fundamental limitation that local synchronization in pairs does not guarantee global consistency across the entire network. Addressing this limitation, our `synchronization base' solution operates on triplets of nodes, providing the crucial advantage that each local synchronization within a triplet inherently implies global synchronization (Appendix~\ref{app:triple_synchronization_proof}. This triplet-based approach not only improves synchronization accuracy but also maintains computational tractability with $O(D^3)$ complexity. 

Building upon the triplet foundation, we then demonstrate the benefits of an `Iterative Refinement' algorithm, which operates on the same triplet structure but leverages the denoised channel estimates $\mH^{'}$ rather than the noisy observations $\mB$. This final experiment demonstrates how our method enables more effective denoising strategies, resulting in superior overall performance in 5G-like scenarios. Each plot includes both a `single channel' and an `ideal line.' These are theoretical results for either not using synchronization at all or assuming perfect synchronization (explained in Appendix~\ref{app:ideal_sync}).

For the experiments, we sample from the ground-truth graphical model and use each method to reconstruct the signal. The problem consists of $6\times 6 = 36$ grids of PRG blocks, with $D = 3\times 4 =12$ measurements made in each block. This means that each block is 3 by 4 measurement cells. The spatial correlation has a length scale of five measurements and follows a squared-exponential decay, which is a simplified model for the Rayleigh-fading scatter~\cite{rayleigh_fading}. Different values of the length scale indicate varying amounts of scatter in the environment and a comparison for multiple values of the length scale is shown in Figure~\ref{fig:heatmap_lengthscale_snr}. 

Figure~\ref{fig:main_sidebyside} shows the main experimental result. It compares all three methods on a 5G MIMO problem. In general, for larger Signal-to-Noise (SNR) ratios, the channel reconstruction achieves lower decibels, and this is reflected in the `ideal synchronization` and `single channel ideal` comparison lines. The right figure redraws each line compared to the `single channel' MSE and indicates the relative improvement. Beyond 0 dB SNR, all three MRA methods exhibit improved performance, with synchronization achieving a lower MSE than pairwise alignment, and Iterative Refinement achieving the lowest MSE across experimental settings.

\vspace{-1mm}
\section{Conclusion} \label{sec:conclusion}
\vspace{-1mm}

Our work presents a significant advancement in multi-reference alignment (MRA) by introducing a novel structured probabilistic framework that addresses the fundamental challenges of MIMO channel denoising. By reformulating the MRA problem in terms of relative poses within a principled statistical model, we eliminate the global redundancies that hinder optimization. We use cycle consistency and arrive at a computationally efficient method that can scale to realistic system sizes.

The iterative refinement algorithm consistently outperforms direct estimation methods and methods that use only pairwise alignment. Across experimental settings, our approach that is decentralized and refines relative pose and channel estimates achieves better reconstruction error consistently. The success of our approach demonstrates the promise of iterative refinement techniques for MIMO synchronization. It opens up new directions for developing more sophisticated denoising strategies that can leverage the global synchronization achieved through our triplet-based cycle consistency.

\bibliographystyle{plainnat}
\bibliography{library}


\newpage
\appendix

\section*{Acknowledgements}
This work is financially supported by Qualcomm Technologies Inc., the University of Amsterdam and the allowance Top consortia for Knowledge and Innovation (TKIs) from the Netherlands Ministry of Economic Affairs and Climate Policy. Qualcomm AI research is an initiative of Qualcomm Technologies, Inc. and/or its subsidiaries. ©2025 Qualcomm Technologies, Inc. and/or its affiliated companies. All Rights Reserved. Correspondence may go to \href{mailto:romijndersrob@gmail.com}{romijndersrob@gmail.com}.

\section{Notation}

\begin{figure}
\centering
\includegraphics[width=0.9\linewidth]{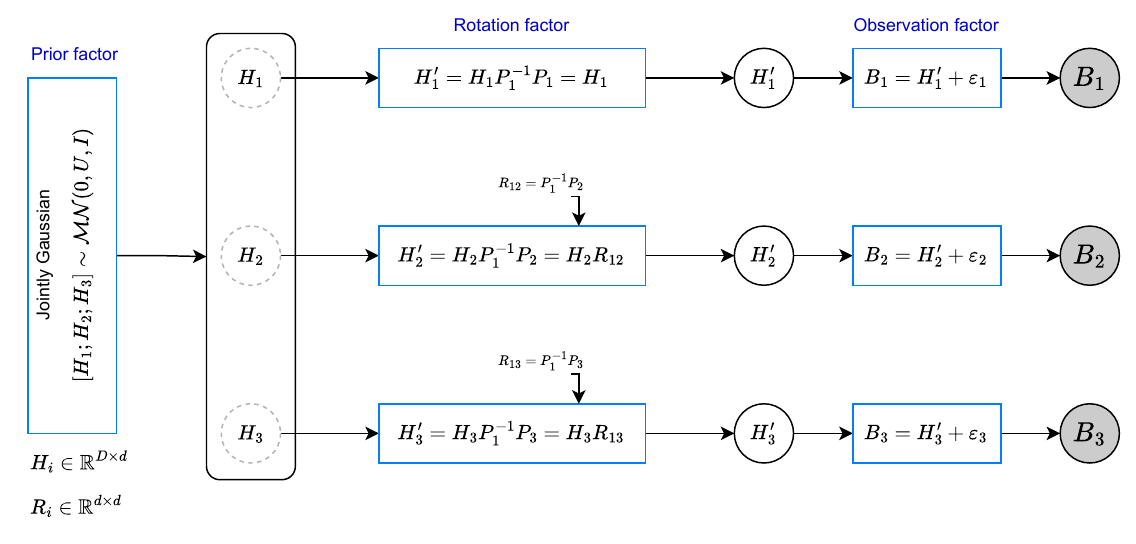}
\caption{Structured Probabilistic Graphical Model to illustrate the joint distribution in Equation~\ref{eqn:joint_distr}}
\label{fig:pgm_pdf}
\end{figure}

An overview of the notation in this paper. Additionally, Figure~\ref{fig:pgm_pdf} presents a factor graph to illustrate how the random variables relate to each other. That factor graph is for three PRG blocks.

\centerline{\bf Notation}
\bgroup
\def\arraystretch{1.5}
\setlength{\tabcolsep}{2pt}
\begin{tabular}{p{1.4in}p{4in}}
$D, d$ & Number of measurements and number of dimensions \\
$H, W$ & Number of rows and columns in the MIMO grid \\
$N = H \cdot W$ & The number of blocks in the MIMO grid \\
\midrule
$\mP_1, \mP_2, \mP_3, \ \dots, \mP_J$ & Rotation matrix of size $\mathbb{R}^{d \times d}$\\
$p(\mP_i)$ & Prior distribution of $\mP_i$, uniform over the rotation angles, $\mP_i \in SO(d)$\\
$\mR_{ij} = \mP_i^{-1}\mP_j$ & Relative rotation matrix of size $\mathbb{R}^{d \times d}$\\
\midrule
$\mH_1, \mH_2, \mH_3, \cdots, \mH_J$ & Hidden matrices of size $\mathbb{R}^{D \times d}$\\
$\mH = [\mH_1; \mH_2; \mH_3]$ & Stacked hidden matrices of size $\mathbb{R}^{3D \times d}$\\
$\mH^{'}_i =  \mH_i \mP_i$ & H-accent is the channel matrix that has undergone an unknown rotation, size $\mathbb{R}^{D \times d}$\\
$\mB_1, \mB_2, \mB_3, \cdots, \mB_J$ & Observed matrices of size $\mathbb{R}^{D \times d}$\\
$\mB = [\mB_1; \mB_2; \mB_3]$ & Stacked observed matrices of size $\mathbb{R}^{3D \times d}$\\
$\varepsilon \sim \mathcal{N}(0, \sigma^2_\varepsilon)$ & Observation noise, shape clear from context \\
$\mB_i = \mH^{'}_i + \varepsilon_i$ & Observation model with iid noise\\
\midrule
$\mathcal{MN}(\mZero, \mU, \mV)$ & Matrix-normal distribution with mean 0, row covariance $\mU$, column covariance $\mV$\\
$p(\mH) = \mathcal{MN}(\mZero, \mU, \mI)$ & Matrix normal distribution with mean 0, row covariance $\mU$\\
\end{tabular}
\egroup
\vspace{0.25cm}

\subsection{Relative poses and channel prior}\label{app:relative_poses}
We model the channel with a Matrix-Normal distribution. The Matrix Normal has the following properties: if $\mX \sim \mathcal{MN}(\mM, \mU, \mV)$ then $\vec{\mX} = \mathcal{N}(\vec{\mM}, \mV \otimes \mU)$. Here, $\vec{\mX}$ is the Frobenius vectorization of a matrix $\mX$ and $\otimes$ is the Kronecker product.

Let's denote the prior as $p(\cdot)$, then the following equivalence holds for the $\mathcal{MN}(0, \mU, \mI)$ matrix normal: 
\begin{align}
p\Big(\begin{bmatrix} \mH_1 \\ \mH_2 \\ \mH_3 \\ \vdots \end{bmatrix} \Big) = p\Big( \begin{bmatrix}
\mH_1 \mP_1^{-1} \\ \mH_2 \mP_1^{-1} \\ \mH_3 \mP_1^{-1} \\ \vdots \end{bmatrix}\Big)  
\end{align}  
This uses the rotated matrices:
\begin{align}
\begin{bmatrix} \mH_1^{'} \\ \mH_2^{'} \\ \mH_3^{'} \\ \vdots \end{bmatrix}
= \begin{bmatrix} \mH_1 \mP_1^{-1}\mP_1 \\ \mH_2 \mP_1^{-1}\mP_2 \\ \mH_3 \mP_1^{-1}\mP_3 \\ \vdots \end{bmatrix} = \begin{bmatrix}
\mH_1  \\ \mH_2 \mR_{12} \\ \mH_3 \mR_{13} \\ \vdots \end{bmatrix} 
\end{align}

\section{Estimate channel using a rotation estimate}

We first derive the estimation of the channel if the rotations were known. We present the case for reconstructing four matrix-valued signals in the presence of relative rotation. The dots, $\cdots$, will indicate generality towards sets of more matrices. The variable $J$ indicates the number of blocks considered in the denoising. For this example, $J=4$.

The derivation starts from $\log p(\mB, \hat{\mR}, \mH)$. This assumes that one has an estimate of the Rotation matrices, $\mR$, and the goal is to find the channel matrices $\mH$ with the largest likelihood.

\small
\begin{align}
&\argmax_{\mH} \log p(\mB, \hat{\mR}, \mH) \\
&=\argmax_{\mH} \log \int_{\mH_1} \cdots \int_{\mH_4} p(\mH_1, \mH_2, \mH_3, \mH_4) p(\mH^{'}_1 | \mH_1) \prod_{j\in [2,3,4]} p(\mH^{'}_j| \mH_j,\hat{\mR}_{1j}) \nonumber \\
& \qquad \qquad \qquad \qquad \qquad \qquad \cdot p(\mB_1|\mH^{'}_1) \big( \prod_{j\in [2,3,4]} p(\mB_j|\mH^{'}_j) \big) d\mH_1 d\mH_2 d\mH_3 d\mH_4\\
&=\argmax_{\mH}= \log \int_{\mH_1} \cdots \int_{\mH_4} p(\mH_1, \mH_2, \mH_3, \mH_4) \delta_{\mH_1 = \mH^{'}_1} \prod_{j\in [2,3,4]} \delta_{\mH_j = \mH^{'}_j\hat{\mR}_{j1}} \nonumber \\
& \qquad \qquad \qquad \qquad \qquad \qquad \cdot \big( \prod_{j\in [1,\cdots,4]} p(\mB_j|\mH^{'}_j) \big) d\mH_1 d\mH_2 d\mH_3 d\mH_4
\end{align}
\normalsize

We integrate over the delta distributions and introduce a change of variables $H^{'}_2 = \mH_2\hat{\mR}_{12}$ and $H^{'}_3 = \mH_3\hat{\mR}_{13}$ and $H^{'}_4 = \mH_4\hat{\mR}_{14}$.

\begin{align}
&\argmax_{\mH} \log  p(\mH^{'}_1, \mH^{'}_2 \hat{\mR}_{21}, \mH^{'}_3 \hat{\mR}_{31}, \mH^{'}_4 \hat{\mR}_{41})   \prod_{j\in [1,\cdots,4]} p(\mB_j|\mH^{'}_j) \\
&=\argmax_{\mH} \log  p(\mH^{'}_1, \mH_2, \mH_3, \mH_4 )  p(\mB_1|\mH^{'}_1) \prod_{j\in [2,3,4]} p(\mB_j|\mH_j\hat{\mR}_{1j})
\end{align}

Next, we write down the actual likelihoods that are Matrix Normal distribution by $\mathcal{MN}$ and align the writing of said distributions $\begin{bmatrix}
\mH_1 \\ \mH_2  \\ \mH_3 \\ \mH_4  \end{bmatrix} \sim \mathcal{MN}(0, \mU, \mI)$.

The operation $\vec{\cdot}$ indicates the ``Frobenius vectorization'' of matrix $\mX$, which concatenates all columns from left to right in one long column vector. This means that we can conveniently write down the Matrix Normal distribution. If $\mX$ has a matrix normal distribution $\mX \sim \mathcal{MN}(0, \mU, \mI)$, then $\vec{\mX}$ has Normal distribution $\vec{\mX} \sim \mathcal{N}(0, \mI \otimes \mU)$ where $\otimes$ indicates the Kronecker product.

\small
\begin{align}
&\argmax_{\mH} \log \mathcal{N} (\vec{\mH^{'}_1,  \mH_2, \mH_3, \mH_4} |  0, \mI_d \otimes \mU) + \log \mathcal{N} (\mH^{'}_1 | \mB_1, \sigma^{2}_\epsilon \mI_{dD}) \nonumber \\ 
& \qquad \qquad + \sum_{j \in [2,3,4]} \log \mathcal{N}(\mH_j | \mB_j \hat{\mR}_{j1}, \sigma^{2}_\epsilon \mI_{dD})  \\
&=\argmax_{\mH} \log \mathcal{N} \Big( \vec{\mH^{'}_1,  \mH_2, \mH_3, \mH_4} |  0, \mI_d \otimes U \Big) \nonumber \\ 
& \qquad \qquad + \log \mathcal{N} \Big( \vec{\mH^{'}_1,  \mH_2, \mH_3, \mH_4} | \vec{\begin{bmatrix}
    \mB_1 \\ \mB_2 \hat{\mR}_{21} \\ \mB_3 \hat{\mR}_{31}  \\ \mB_4 \hat{\mR}_{41}
    \end{bmatrix}}, \mI_d \otimes \sigma^{2}_\epsilon \mI_{4D} \Big) \\
&=\argmax_{\mH} \log \mathcal{N} \Big( \vec{\mH^{'}_1,  \mH_2, \mH_3, \mH_4} \  | \ \mSigma_{\mH}  \vec{\sigma^{-2}_\epsilon \mB_1, \sigma^{-2}_\epsilon \mB_2 \hat{\mR}_{21}, \sigma^{-2}_\epsilon \mB_3 \hat{\mR}_{31}, \sigma^{-2}_\epsilon \mB_4 \hat{\mR}_{41}} , \ \mSigma_{\mH} \Big)
\end{align}
\normalsize

We use shorthand for the denoising correlation matrix:
\begin{align}
\mSigma_{\mH} &= \Big((\mI_d \otimes \mU)^{-1} + (\mI_d \otimes \sigma^{2}_\epsilon \mI_{JD})^{-1}\Big)^{-1}  \nonumber \\
&= \Big((\mI_d \otimes \mU)^{-1} + \sigma^{-2}_\epsilon \mI_{JDd}\Big)^{-1}  \nonumber \\
&= (\begin{bmatrix}
        (\mU^{-1} + \sigma^{-2}_\epsilon\mI_{JD})^{-1} & 0 & 0 & \cdots \\
        0 & (\mU^{-1} + \sigma^{-2}_\epsilon \mI_{JD})^{-1} & 0 & \cdots \\
        0 & 0 & (\mU^{-1} + \sigma^{-2}_\epsilon \mI_{JD})^{-1}& \cdots \\
        \vdots & \vdots & \vdots & \ddots \end{bmatrix} )^{-1} \nonumber \\
&= \mI_d \otimes \big( \mU^{-1} + \sigma^{-2}_\epsilon \mI_{4D} \big)^{-1}.
\end{align}
\normalsize

The argmax of a Gaussian likelihood is the mean parameter of the said Gaussian distribution. Therefore, we find the mean parameter of $H^{'}$ from:
\begin{align}
\begin{bmatrix} \mH^{'}_{1, t+1} \\ \mH^{'}_{2, t+1}\hat{\mR}_{21} \\ \mH^{'}_{3, t+1}\hat{\mR}_{31} \\ \mH^{'}_{4, t+1}\hat{\mR}_{41} \end{bmatrix} &=  (\mI_d \otimes (\Big(\mU^{-1} + \sigma^{-2}_\epsilon \mI_{4D}\Big)^{-1} ) )
\begin{bmatrix} \sigma^{-2}_\epsilon \mB_1 \\ \sigma^{-2}_\epsilon \mB_2 \hat{\mR}_{21} \\ \sigma^{-2}_\epsilon \mB_3 \hat{\mR}_{31} \\ \sigma^{-2}_\epsilon \mB_4 \hat{\mR}_{41}\end{bmatrix}. \label{eqn:m-step__meanH_fourleaves}
\end{align}

The general equation for general $J$ would be:
\begin{align}
\begin{bmatrix} \mH^{'}_{1, t+1} \\ \mH^{'}_{2, t+1}\hat{\mR}_{21} \\ \mH^{'}_{3, t+1}\hat{\mR}_{31} \\ \mH^{'}_{4, t+1}\hat{\mR}_{41}  \\ \vdots \end{bmatrix} &=  (\mI_d \otimes (\Big(\mU^{-1} + \sigma^{-2}_\epsilon \mI_{JD}\Big)^{-1} ) )
\begin{bmatrix} \sigma^{-2}_\epsilon \mB_1 \\ \sigma^{-2}_\epsilon \mB_2 \hat{\mR}_{21} \\ \sigma^{-2}_\epsilon \mB_3 \hat{\mR}_{31} \\ \sigma^{-2}_\epsilon \mB_4 \hat{\mR}_{41} \\ \vdots \end{bmatrix}. \label{eqn:m-step__meanH_Jleaves}
\end{align}

\section{Directly estimate rotations}\label{app:direct3_derivation}

For the direct approach, we estimate the rotation matrices within each subgraph. This step occurs even before iterative refinement and provides the first estimate of the rotation matrices based on the noisy observations.

\subsection{MLE, three obs, directly from B}
The comments prefixed with \textcolor{blue}{`...'} in \textcolor{blue}{blue} are comments to explain the derivation steps.

\scriptsize
\begin{align}
& \hat{\mR}_{12}, \hat{\mR}_{13} \nonumber \\
&= \argmax_{\mR_{12},\mR_{13}} \ \log p(\mB_1, \mB_2, \mB_3 | \mR_{12},\mR_{13}) \label{eqn:direct3_opening}\\
& \textcolor{blue}{\text{... Introduce and marginalize over $H_1$, $H_2$, and $H_3$}} \nonumber \\
&= \argmax_{\mR_{12},\mR_{13}} \ \log \int_{\mH_1} \int_{\mH_2} \int_{\mH_3} p(\mH_1, \mH_2, \mH_3) \int_{\mH^{'}_1} \int_{\mH^{'}_2} \int_{\mH^{'}_3} p(\mB_1 | \mH^{'}_1) p(\mB_2 | \mH^{'}_2) p(\mB_3 | \mH^{'}_3) \nonumber \\
&\qquad \qquad \qquad p({\mH^{'}_1} | \mH_1) \ p({\mH^{'}_2}|\mH_2,\mR_{12}) \ p({\mH^{'}_3}|\mH_3,\mR_{13}) \ d\mH^{'}_1  d\mH^{'}_2  d\mH^{'}_3 \ d\mH_1  d\mH_2  d\mH_3  \\
&= \argmax_{\mR_{12},\mR_{13}} \ \log \int_{\mH_1} \int_{\mH_2} \int_{\mH_3} p(\mH_1, \mH_2, \mH_3) \int_{\mH^{'}_1} \int_{\mH^{'}_2} \int_{\mH^{'}_3} p(\mB_1 | \mH^{'}_1) p(\mB_2 | \mH^{'}_2) p(\mB_3 | \mH^{'}_3) \nonumber \\
&\qquad \qquad \qquad \delta_{\mH_1 = {\mH^{'}_1}} \ \delta_{ \mH_2 = {\mH^{'}_2}\mR_{21}} \ \delta_{\mH_3 = {\mH^{'}_3}\mR_{31}} \ d\mH^{'}_1  d\mH^{'}_2  d\mH^{'}_3 \ d\mH_1  d\mH_2  d\mH_3  \\
& \textcolor{blue}{\text{... Integrate against delta distributions}} \nonumber \\
&= \argmax_{\mR_{12},\mR_{13}} \ \log   \int_{\mH^{'}_1} \int_{\mH^{'}_2} \int_{\mH^{'}_3}  p({\mH^{'}_1}, {\mH^{'}_2}\mR_{21}, {\mH^{'}_3}\mR_{31}) \ p(\mB_1 | \mH^{'}_1) p(\mB_2 | \mH^{'}_2) p(\mB_3 | \mH^{'}_3) \  d\mH^{'}_1  d\mH^{'}_2  d\mH^{'}_3   \\
& \textcolor{blue}{\text{... Make Matrix Normal explicit}} \nonumber \\
&= \argmax_{\mR_{12},\mR_{13}} \ \log   \int_{\mH^{'}_1} \int_{\mH^{'}_2} \int_{\mH^{'}_3}  \mathcal{MN}(\begin{bmatrix} {\mH^{'}_1} \\ {\mH^{'}_2}\mR_{21} \\ {\mH^{'}_3}\mR_{31} \end{bmatrix}; 0, \mU, \mI_d) \ \mathcal{MN}(\mH^{'}_1 | \mB_1, \sigma^2 \mI_D, \mI_d)   \nonumber \\
& \qquad \qquad \qquad \mathcal{MN}({\mH^{'}_2}\mR_{21} ; {B_2}\mR_{21}, \sigma^2 \mI_D, \mI_d) \  \mathcal{MN}({\mH^{'}_3}\mR_{31} ; {B_3}\mR_{31}, \sigma^2 \mI_D, \mI_d) \  d\mH^{'}_1  d\mH^{'}_2  d\mH^{'}_3   \\
& \textcolor{blue}{\text{... Fuse Matrix normals}} \nonumber \\
&= \argmax_{\mR_{12},\mR_{13}} \ \log   \int_{\mH^{'}_1} \int_{\mH^{'}_2} \int_{\mH^{'}_3}  \mathcal{MN}(\begin{bmatrix} {\mH^{'}_1} \\ {\mH^{'}_2}\mR_{21} \\ {\mH^{'}_3}\mR_{31} \end{bmatrix}; 0, \mU, \mI_d) \ \mathcal{MN}(\begin{bmatrix} {\mH^{'}_1} \\ {\mH^{'}_2}\mR_{21} \\ {\mH^{'}_3}\mR_{31} \end{bmatrix} | \begin{bmatrix} \mB_1 \\ \mB_2R_{21} \\ \mB_3 \mR_{31} \end{bmatrix}, \sigma^2 \mI_{3D}, \mI_d) \  d\mH^{'}_1  d\mH^{'}_2  d\mH^{'}_3  \\
\nonumber \\
&= \argmax_{\mR_{12},\mR_{13}} \ \log     \mathcal{MN}(\begin{bmatrix} \mB_1 \\ \mB_2R_{21} \\ \mB_3 \mR_{31} \end{bmatrix}; 0, \mU + \sigma^2 I, \mI_d)
\end{align}
\normalsize

The simplification from two to one Matrix Normal uses some tricks: the multiplication of two normal distributions yields a normal distribution, and the integral will not change under rotation. Therefore, we can use standard properties of the Matrix Normal product. The argmax over the matrix normal can be obtained by writing out the Trace and removing scalar constants:

\begin{align}
& \hat{\mR}_{12}, \hat{\mR}_{13} \nonumber \\
&=\argmax_{\mR_{12},\mR_{13}} \frac{-1}{2} \Tr  \Big[ \begin{bmatrix}
       \mB_1 \\ \mB_2R_{21} \\ \mB_3 \mR_{31} \end{bmatrix}^T ( \mU+\sigma^2 \mI)^{-1} \begin{bmatrix}
       \mB_1 \\ \mB_2R_{21} \\ \mB_3 \mR_{31} \end{bmatrix} \Big] \nonumber \\ 
&= \argmax_{\mR_{12},\mR_{13}} \frac{-1}{2} \Tr  \Big[ \begin{bmatrix}
       \mB_1 \\ \mB_2R_{21} \\ \mB_3 \mR_{31} \end{bmatrix} \begin{bmatrix}
       \mB_1 \\ \mB_2R_{21} \\ \mB_3 \mR_{31} \end{bmatrix}^T ( \mU+\sigma^2\mI)^{-1}  \Big] \\
&= \argmax_{\mR_{12},\mR_{13}}  -\Tr  \Big[ \begin{bmatrix}
B_1 \mB_1^T & \mB_1 \mR_{12} \mB_2^T & \mB_1 \mR_{13} \mB_3^T \\
B_2 \mR_{21} \mB_1^T & \mB_2 \mB_2^T & \mB_2 \mR_{21} \mR_{13} \mB_3^T \\
B_3 \mR_{31}  \mB_1^T & \mB_3 \mR_{31} \mR_{12} \mB_2^T & \mB_3 \mB_3^T \end{bmatrix} ( \mU+\sigma^2\mI)^{-1}  \Big] \label{eqn:B_direct_before_split_in_U}
\end{align}

Equation~\ref{eqn:B_direct_before_split_in_U} considers the matrix as composed of submatrices. Since the $U$ matrix is always composed of $J \cdot D$ by $J \cdot D$ elements, we can always split it in $J$ by $J$ submatrices of size $D \times D$:

\begin{equation}\label{eqn:subslice_mU}
-( \mU+\sigma^2\mI)^{-1} =    \begin{bmatrix}
\mU_{1} & \mU_{a} & \mU_{b} \\ \mU^T_{a} & \mU_{2} & \mU_c  \\ \mU^T_{b} & \mU^T_{c} & \mU_3
    \end{bmatrix}
\end{equation}

\begin{align}
\hat{\mR}_{12}, \hat{\mR}_{13} &= \argmax_{\mR_{12},\mR_{13}}  \Tr  \Big[ \begin{bmatrix}
       \mB_1 \mB_1^T & \mB_1 \mR_{12} \mB_2^T & \mB_1 \mR_{13} \mB_3^T \\
       \mB_2 \mR_{21} \mB_1^T & \mB_2 \mB_2^T & \mB_2 \mR_{21} \mR_{13} \mB_3^T \\
       \mB_3 \mR_{31}  \mB_1^T & \mB_3 \mR_{31} \mR_{12} \mB_2^T & \mB_3 \mB_3^T \end{bmatrix} \begin{bmatrix}
    \mU_{1} & \mU_{a} & \mU_{b} \\ \mU^T_{a} & \mU_{2} & \mU_c  \\ \mU^T_{b} & \mU^T_{c} & \mU_3
    \end{bmatrix}  \Big] \\
& \textcolor{blue}{\text{...Split to individual traces}} \nonumber \\
    &= \argmax_{\mR_{12},\mR_{13}}  2\Tr \Big[  \mB_1 \mR_{12} \mB_2^T \mU_a^T \Big] + 2\Tr \Big[  \mB_1 \mR_{13} \mB_3^T \mU_b^T \Big]  + 2\Tr \Big[  \mB_2 \mR_{21} \mR_{13} \mB_3^T \mU_c^T \Big] \label{eqn:B_direct_before_text_procrustus}
\end{align}
The argument in Equation~\ref{eqn:B_direct_before_text_procrustus} is linear in both rotation matrices. In each argument, there is a closed-form expression for the optimum over the set of orthogonal matrices. Therefore, we suggest to iteratively determine this optimum. With each closed form solution, each step is guaranteed to non-decrease the objective value and we find that this usually converges to machine precision in less than eight steps.

\begin{equation}
\text{iterate}
\begin{cases}
\mR_{13} = \argmax  \langle  \mR_{21}, \mB_2^T  \mU_a^T \mB_1 \rangle_F + \langle  \mR_{31}, \mB_3^T  \mU_b^T \mB_1 \rangle_F  + \langle  \mR_{31}, \mB_3^T  \mU_c^T \mB_2 \mR_{21} \rangle_F  \\
\mR_{12} = \argmax  \langle  \mR_{21}, \mB_2^T  \mU_a^T \mB_1 \rangle_F + \langle  \mR_{31}, \mB_3^T  \mU_b^T \mB_1 \rangle_F  + \langle  \mR_{21}, \mB_2^T  \mU_c \mB_3 \mR_{31} \rangle_F
\end{cases}
\end{equation}

The closed-form solution is known as the Orthogonal Procrustes problem~\cite{procrustes01,procrustes02}. The projection is achieved by removing the singular values from the Singular-Value Decomposition:

\begin{align}
\argmax_\mR \langle  \mR, \mX \rangle_F  = \mA L \mV^T
\end{align}
from Singular Value Decomposition $\mX=\mA\Sigma\mV^T$ and with $L=I$ if $\det(\mX)>0$ or $L=\text{diag}(1, 1, 1, \dots, -1)$ otherwise (assuming the smallest singular-value of $X$ is at the bottom of $\Sigma$).

\section{Iterative refinement 3}\label{app:iterative3_derivation}

Deriving the iterative refinement looks much like Equation~\ref{eqn:direct3_opening}. However, it crucially starts from optimizing the rotation matrices given the channel estimates $\mH$, instead of the noisy estimates $\mB$.

The comments prefixed with \textcolor{blue}{`...'} are comments to explain the derivation steps.

\scriptsize
\begin{align}
& \hat{\mR}_{12}, \hat{\mR}_{13} \nonumber \\
&= \argmax_{\mR_{12},\mR_{13}} \ \log p(\mR_{12},\mR_{13}|\mH^{'}) \\
&= \argmax_{\mR_{12},\mR_{13}} \ \log p(\mH^{'}_{1}, \mH^{'}_{2}, \mH^{'}_{3} | \mR_{12},\mR_{13}) \\
& \textcolor{blue}{\text{... Introduce and marginalize over $H_1$, $H_2$, and $H_3$}} \nonumber \\
&= \argmax_{\mR_{12},\mR_{13}} \ \log \int_{\mH_1} \int_{\mH_2} \int_{\mH_3} p(\mH_1, \mH_2, \mH_3) \ p(\mH^{'}_1 | \mH_1) \ p(\mH^{'}_2|\mH_2,\mR_{12}) \ p(\mH^{'}_3|\mH_3,\mR_{13}) \ d\mH_1 \ d\mH_2 \ d\mH_3\\
&= \argmax_{\mR_{12},\mR_{13}} \  \log \int_{\mH_1} \int_{\mH_2} \int_{\mH_3} p(\mH_1, \mH_2, \mH_3) \ \delta_{\mH_1 = \mH^{'}_1} \ \delta_{ \mH_2 = \mH^{'}_2\mR_{21}} \ \delta_{ \mH_3 = \mH^{'}_3\mR_{31}} \ d\mH_1 \ d\mH_2 \ d\mH_3\\
&= \argmax_{\mR_{12},\mR_{13}} \log p(\mH^{'}_1, \mH^{'}_2\mR_{21}, \mH^{'}_3\mR_{31}) 
\end{align}
\normalsize

We'll write down the matrix-normal form of the channel prior: $\begin{bmatrix}
        \mH_1 \\ \mH_2  \\ \mH_3 \end{bmatrix} \sim \mathcal{MN}(0, \mU, \mI)$:

\begin{align}
& \hat{\mR}_{12}, \hat{\mR}_{13} \nonumber \\
&=\argmax_{\mR_{12},\mR_{13}} \frac{-1}{2} \Tr  \Big[ \begin{bmatrix}
       \mH^{'}_1 \\ \mH^{'}_2\mR_{21} \\ \mH^{'}_3\mR_{31} \end{bmatrix}^T \mU^{-1} \begin{bmatrix}
       \mH^{'}_1 \\ \mH^{'}_2\mR_{21} \\ \mH^{'}_3\mR_{31} \end{bmatrix} \Big] \nonumber \\ 
&= \argmax_{\mR_{12},\mR_{13}} \frac{-1}{2} \Tr  \Big[ \begin{bmatrix}
       \mH^{'}_1 \\ \mH^{'}_2\mR_{21} \\ \mH^{'}_3\mR_{31} \end{bmatrix} \begin{bmatrix}
       \mH^{'}_1 \\ \mH^{'}_2\mR_{21} \\ \mH^{'}_3\mR_{31} \end{bmatrix}^T \mU^{-1}  \Big] \\
&= \argmax_{\mR_{12},\mR_{13}}  -\Tr  \Big[ \begin{bmatrix}
       \mH^{'}_1 \mH^{'T}_1 & \mH^{'}_1 \mR_{12} \mH^{'T}_2 & \mH^{'}_1 \mR_{13} \mH^{'T}_3 \\
       \mH^{'}_2 \mR_{21} \mH^{'T}_1 & \mH^{'}_2 \mH^{'T}_2 & \mH^{'}_2 \mR_{21} \mR_{13} \mH^{'T}_3 \\
       \mH^{'}_3 \mR_{31}  \mH^{'T}_1 & \mH^{'}_3 \mR_{31} \mR_{12} \mH^{'T}_2 & \mH^{'}_3 \mH^{'T}_3 \end{bmatrix} \mU^{-1}  \Big] \\
& \textcolor{blue}{\text{...Consider $-\mU^{-1}$ a block matrix $\begin{bmatrix}
        \mU_{1} & \mU_{a} & \mU_{b} \\ \mU^T_{a} & \mU_{2} & \mU_{c}  \\ \mU^T_{b} & \mU^T_{c} & \mU_{3}
    \end{bmatrix}$ with $\mU \in \mathbb{\mR}^{3D \times 3D}$ and $\mU_a \in \mathbb{\mR}^{D \times D}$}} \nonumber \\
&= \argmax_{\mR_{12},\mR_{13}}  \Tr  \Big[ \begin{bmatrix}
       \mH^{'}_1 \mH^{'T}_1 & \mH^{'}_1 \mR_{12} \mH^{'T}_2 & \mH^{'}_1 \mR_{13} \mH^{'T}_3 \\
       \mH^{'}_2 \mR_{21} \mH^{'T}_1 & \mH^{'}_2 \mH^{'T}_2 & \mH^{'}_2 \mR_{21} \mR_{13} \mH^{'T}_3 \\
       \mH^{'}_3 \mR_{31}  \mH^{'T}_1 & \mH^{'}_3 \mR_{31} \mR_{12} \mH^{'T}_2 & \mH^{'}_3 \mH^{'T}_3 \end{bmatrix} \begin{bmatrix}
    \mU_{1} & \mU_{a} & \mU_{b} \\ \mU^T_{a} & \mU_{2} & \mU_{c}  \\ \mU^T_{b} & \mU^T_{c} & \mU_{3}
    \end{bmatrix}  \Big] \\
& \textcolor{blue}{\text{...Split to individual traces}} \nonumber \\
    &= \argmax_{\mR_{12}, \mR_{13}}  2\Tr \Big[  \mH^{'}_1 \mR_{12} \mH^{'T}_2 \mU_a^T \Big] + 2\Tr \Big[  \mH^{'}_1 \mR_{13} \mH^{'T}_3 \mU_b^T \Big]  + 2\Tr \Big[  \mH^{'}_2 \mR_{21} \mR_{13} \mH^{'T}_3 \mU_c^T \Big] \label{eqn:iter3_before_text_procrustus}
\end{align}

The optimization argument in Equation~\ref{eqn:iter3_before_text_procrustus} is linear in both rotation matrices. In each argument, there is a closed-form expression for the optimum over the set of orthogonal matrices. Therefore, we suggest to iteratively determine this optimum. With each closed form solution, each step is guaranteed to non-decrease the objective value and we find that this usually converges to machine precision in less than eight steps.

\begin{align}
&\text{iterate until convergence} \nonumber \\
&\begin{cases}
\mR_{12} = \argmax_{\mR_{12}}  \langle  \mR_{21}, \mH^{'T}_2  \mU_a^T \mH^{'}_1 \rangle_F + \textcolor{gray}{\langle  \mR_{31}, \mH^{'T}_3  \mU_b^T \mH^{'}_1 \rangle_F}  + \langle  \mR_{21}, \mH^{'T}_2  \mU_c \mH^{'}_3 \mR_{31} \rangle_F  \\
\mR_{13} = \argmax_{\mR_{13}}  \textcolor{gray}{\langle  \mR_{21}, \mH^{'T}_2  \mU_a^T \mH^{'}_1 \rangle_F} + \langle  \mR_{31}, \mH^{'T}_3  \mU_b^T \mH^{'}_1 \rangle_F  + \langle  \mR_{31}, \mH^{'T}_3  \mU_c^T \mH^{'}_2 \mR_{21} \rangle_F \nonumber
\end{cases}
\end{align}
\normalsize
The closed-form solution to the latter argmax is the same as the Procrustes problem in Section~\ref{app:direct3_derivation}. Each step is a closed-form maximizer, so the iteration is guaranteed to converge.

\section{Local to Global synchronization}\label{app:triple_synchronization_proof}

This section includes the theoretical proofs behind our main arguments for reducing the absolute pose estimation problem to a relative pose synchronization problem.
In particular, we will show 
\begin{itemize}
    \item[\emph{1.}] that the set of absolute poses and the set of cycle-consistent relative poses are in one-to-one correspondence (up to a global transformation of the absolute poses) and that
    \item[\emph{2.}] if all $3$-cycles (triplets) in a surface mesh (e.g., the physical resource blocks in the frequency-time lattice in the 5G NR channel) are cycle-consistent, then the whole mesh is cycle consistent.
\end{itemize}

First, we define the notion of cycle consistency precisely.
Consider a \textbf{connected graph} $G=(V, E)$ with a set $V$ of $N$ nodes and a set $E$ of edges. 
Then, an assignment of relative poses $R=\{R_{ij}\}_{(i,j)\in E}$ is cycle consistent along a $k$-cycle $\gamma = [\gamma_1, \gamma_2, \dots, \gamma_k, \gamma_1]$ (with $\gamma_i \in V$ and $(\gamma_i, \gamma_{i+1}) \in E$) if
\begin{align}
    R_{\gamma_1,\gamma_2}
    R_{\gamma_2,\gamma_3}
    \cdots
    R_{\gamma_k,\gamma_1} = I
\end{align}
An assignment of relative poses $\{R_{ij}\}_{(i,j)\in E}$ is cycle consistent if it is cycle-consistent over all cycles in the graph $G$.

\subsection{Absolute to relative poses}\label{app:absolute_to_relative}

We need to prove two directions. First, any set of absolute poses $P=\{P_i\}_{i=1}^N$ is associated with one unique set of cycle-consistent relative poses $R=\{R_{ij}\}_{1\leq i,j\leq N}$.
This true by definition, since $R_{ij} := P_i^{-1}P_j$ and for any cycle $\gamma = [\gamma_1, \gamma_2, \dots, \gamma_k, \gamma_1]$:
\begin{align}
 R_{\gamma_1,\gamma_2} R_{\gamma_2,\gamma_3} \cdots R_{\gamma_k,\gamma_1} =& P_{\gamma_1}^{-1}P_{\gamma_2} P_{\gamma_2}^{-1}P_{\gamma_3} \cdots P_{\gamma_k}P_{\gamma_1} = P_{\gamma_1}^{-1}P_{\gamma_1} = I
\end{align}

Second, we show that any set of cycle-consistent relative poses corresponds to an \emph{equivalence class} of absolute poses (defined up to a global transformation).
First, we note that cycle-consistency implies that, for any pair of nodes $i, j$, walking along any path connecting them generates the same transformation, i.e., the relative pose $R_{ij}$ is well defined (even if edge $(i, j) \notin E$).
For any pair of paths $\alpha, \beta$ of length $a$ and $b$ connecting $i$ to $j$, i.e. $\alpha_1=\beta_1=i$ and $\alpha_{a}=\beta_{b}=j$, concatenating $\alpha$ and $\beta$ reversed form a cycle $\gamma = [\alpha_1=i, \alpha_2, \alpha_3, \dots, \alpha_{a}= j =\beta_{b}, \dots, \beta_3, \beta_2, \beta_1=i]$.
Since the relative poses $R$ are cycle-consistent, it follows that
\begin{align}
    R_{\alpha_1=i,\alpha_2}
    R_{\alpha_2,\alpha_3}
    \cdots
    R_{\alpha_{a-1},\alpha_a=j}
    R_{\beta_{b}=j,\beta_{b-1}}
    \cdots
    R_{\beta_2,\beta_1=i} &= I \\
    R_{\alpha_1=i,\alpha_2}
    R_{\alpha_2,\alpha_3}
    \cdots
    R_{\alpha_{a-1},\alpha_a=j} 
    &=
    R_{\beta_2,\beta_1=i}^{-1}
    \cdots
    R_{\beta_{b}=j,\beta_{b-1}}^{-1} \\
    R_{\alpha_1=i,\alpha_2}
    R_{\alpha_2,\alpha_3}
    \cdots
    R_{\alpha_{a-1},\alpha_a=j} 
    &=
    R_{\beta_1=i,\beta_2}
    \cdots
    R_{\beta_{b-1}, \beta_{b}=j}
     =: R_{i,j}
\end{align}

Hence, one can construct a set of absolute poses $P$ by choosing an arbitrary pose for the first node $P_0$ and then defining $P_i := P_0 R_{0,i}$.
Note that any choice of $P_0$ is valid, but leads to a different set of absolute poses.
In particular, any two such equivalent sets of absolute poses $P$ and $P'$ differ precisely by a global transformation, i.e., $\exists T : \forall i, P_i'=T P_i$.
Moreover, picking a different starting node $j$ rather than $0$ results in the equivalent solution one would have obtained by instead picking $P_0 = T R_{j,0}$.

Finally, any of these equivalent solutions generates the initial set of relative poses; hence, we have proved the bi-directional correspondence.

\subsection{Sufficiency of $3$-cycles}

Consider a graph $G=(V, E)$ representing the edges of a triangulated \textbf{planar} mesh, i.e. each node in $V$ can be embedded in $\mathbb{R}^2$ with no edges crossing.
Given a certain assignment of relative poses $R$ on $G$, if all $3$-cycles in $G$ (i.e. all triangles in the mesh) are cycle-consistent, then any cycle $\gamma$ in $G$ is also cycle-consistent, i.e.:
\begin{align}
    R_\gamma = 
    R_{\gamma_1,\gamma_2}
    R_{\gamma_2,\gamma_3}
    \cdots
    R_{\gamma_k,\gamma_1} = I
\end{align}

To prove this, note that a cycle $\gamma$ is the boundary of an enclosed area $\mathcal{U}_\gamma \subset \mathbb{R}^2$ which includes a subset $V_\gamma \subset V$ of nodes and a subset $E_\gamma \subset E$ of edges (including those in $\gamma$).
We prove that $R_\gamma = I$ \emph{by induction} on the size of $E_\gamma$.

\emph{Base case} $|E_\gamma| = 3$: this implies the cycle $\gamma$ contains a single triangle and, therefore, it's already cycle-consistent by assumption, i.e. $R_\gamma = I$.

\emph{Inductive case} $|E_\gamma| > 3$: consider any edge $(i,j) \in \gamma$. There exists a node $k \in V_\gamma$ such that $(i, k)$ and $(j, k) \in E_\gamma$.
Then, $(i, j, k)$ for a cycle-consistent triangle.
Now we consider three cases.
\begin{itemize}
    \item If $(j, k) \in \gamma$, we can define a new cycle $\gamma'$ by replacing both edges $(i, j)$ and $(j, k)$ with the edge $(i, k)$. Note that $R_\gamma = R_{\gamma'}$ since $R_{ij} R_{jk} = R_{ik}$, but $E_{\gamma'} = E_\gamma \cup \{(i, j), (j, k)\}$ and, therefore, $E_{\gamma'} = E_\gamma -2$.
    \item If $(i, k) \in \gamma$, one can apply a similar construction.
    \item Otherwise, we can define a new cycle $\gamma'$ by replacing the edge $(i, j)$ with the two edge $(i, k)$ and $(k, j)$. Note that $R_\gamma = R_{\gamma'}$ since $R_{ij} = R_{ik} R_{jk}^{-1}$, but $E_{\gamma'} = E_\gamma \cup \{(i, j)\}$ and, therefore, $E_{\gamma'} = E_\gamma -1$.
\end{itemize}%
Thus, the relative poses $R$ are cycle-consistent on $\gamma$.

More generally, the set of all sufficient cycles for a generic graph is the \emph{cycle basis} for that graph.
When considering meshes over more general surfaces, the result proved above is not necessarily true, i.e. even if all triangles are cycle consistent, there might exist cycles which are not.
These cycles can be precisely identified in the first homology group of the surface.
For instance, a $2$-torus has two additional cycles, but a similar proof works for a spherical surface.

\section{Performance under ideal synchronization}\label{app:ideal_sync}

For many of the results, we are seeking a comparison for the empirical results. Without the precoding, the MIMO task would amount to reconstructing a signal under structured Gaussian noise. As such, we can characterize the Mean-Squared Error under `ideal synchronization' in closed form. This Appendix section aims to characterize the MSE value so that we can plot it in the diagrams.

The `ideal synchronization' MSE is instructive to have, but it is not a strict lower bound, however. Due to finite sample sizes, some reconstructions could achieve lower MSE. Conversely, due to finite sample sizes, the performance might not be attainable. Therefore, it shouldn't be considered a `target performance' either. The latter is due to the interaction between rotation and signal noise: two signals could be noised such that the optimal rotation on a finite sample under some probabilistic model would not match the true rotation.

In general, the case where rotations are known could be modeled as the reconstruction of a correlated Gaussian distribution. Consider the following system. Matrix $\mX$ is drawn from a correlated Gaussian distribution. After additive i.i.d. Gaussian noise on each of the elements in the matrix, a noisy matrix $\mY$ is observed. Then, we aim to characterize the reconstruction error $\mathbb{E}[\|X - \hat{\mX}(\mY)\|^2]$.
\begin{align}
\mX \sim \gN(0, \mSigma_X); \qquad  \epsilon \sim \gN(0, \mI); \qquad \mY = \mX + \sigma \epsilon
\end{align}

We are interested in the expected MSE of estimating $\mX$ from $\mY$, as a function of $\mSigma_X$ and $\sigma$.
\begin{align}
    \mathcal{L}(\mSigma_X, \sigma) = \mathbb{E}[\|X - \hat{\mX}(\mY)\|^2] &= \mathbb{E}[X^TX + \hat{\mX}(\mY)^T\hat{\mX}(\mY) - 2 X^T\hat{\mX}(\mY)]
\end{align}

The covariance matrix $\mSigma_X$ is of size $ND$ by $ND$ elements, which is the covariance between all measurements in all blocks.
The optimal estimator is the MMSE estimator $hat{\mX}(\mY) = \mathbb{E}[X | Y]$. This follows from the conditional distribution: $p(X|Y) \propto p(Y|X)p(X) \propto \gN(\mu_c, \mSigma_c)$.

This follows from estimating $X$ with the MMSE estimator:
\begin{align}
\mSigma_c &= \left(\mSigma_X^{-1} + \frac{1}{\sigma^2} \mI \right)^{-1} = \sigma^2 \left(\sigma^2\mSigma_X^{-1} + \mI \right)^{-1}  \\
\mmu_c &= \mSigma_c \left(\frac{1}{\sigma^2} \mY \right) \\
\hat{\mX}(\mY) &= \left( \mSigma_X^{-1} + \frac{1}{\sigma^2} \mI \right)^{-1} \left( \frac{1}{\sigma^2} \mY \right) = \left( \sigma^2 \mSigma_X^{-1} + \mI \right)^{-1} \mY \label{eqn:mmse-estimator-correlated-noised-gaussian}
\end{align}

For the derivation, we will need the second moment matrices: $\mathbb{E}[\mX\mX^T] = \mSigma_X, \ \mathbb{E}[\mY\mY^T] = \mSigma_X + \sigma^2 \mI, \ \mathbb{E}[\mY\mX^T] = \mSigma_X$.

\textbf{Linear MMSE} 
The covariance of the linear MMSE estimator is given by $\mSigma_X - \mSigma_{XY} \mSigma_Y^{-1} \mSigma_{YX}$. This follows from searching over the space of linear estimators: $\hat{\mX}(\mY) = \mW\mY + b$. The minimal error is achieved when $\mW = \mSigma_{XY}\mSigma_\mY^{-1}$. Therefore, the MMSE is 
\begin{align}\label{eqn:ideal_synchro_mmse}
    \hat{\mathcal{L}}(\mSigma_X, \sigma) &= \mSigma_X - \mSigma_{X} (\Sigma_X + \sigma^2 I)^{-1} \mSigma_{X}
\end{align}



Equation~\ref{eqn:ideal_synchro_mmse} can also be used to characterize the performance of the per-block MSE. That is, the performance of considering any rotations at all and treating each PRG block in isolation. In that case, $\mSigma_X \in \mathbb{R}^{D \times D}$ is the correlated matrix of all measurements within one cell, instead of $\mSigma_X \in \mathbb{R}^{JD \times JD}$.

\end{document}